\documentclass[sigconf]{acmart}

\usepackage{graphicx}
\usepackage{booktabs}
\usepackage{subcaption}

\AtBeginDocument{%
  }

\copyrightyear{2026}
\acmYear{2026}
\setcopyright{cc}
\setcctype{by}
\acmConference[ICMI Companion '26]{Companion of the INTERNATIONAL CONFERENCE ON MULTIMODAL INTERACTION}{October 05--09, 2026}{Napoli, Italy}
\acmBooktitle{Companion of the INTERNATIONAL CONFERENCE ON MULTIMODAL INTERACTION (ICMI Companion '26), October 05--09, 2026, Napoli, Italy}
\acmDOI{10.1145/3776591.3832513}
\acmISBN{979-8-4007-2319-3/2026/10}

\begin{document}

%%
%% The "title" command has an optional parameter,
%% allowing the author to define a "short title" to be used in page headers.
\title[Context-Aware Prefaces for Low-Latency Turn-Taking]{Low-Latency Turn-Taking via Context-Aware Preface Generation in a Real-World Dialogue Robot}

%%
%% The "author" command and its associated commands are used to define
%% the authors and their affiliations.
%% Of note is the shared affiliation of the first two authors, and the
%% "authornote" and "authornotemark" commands
%% used to denote shared contribution to the research.
\author{Yuki Okafuji}
\correspondingauthor
\orcid{0000-0002-9547-3681}
\affiliation{%
  \institution{CyberAgent}
  \city{Tokyo}
  \country{Japan}
}
\affiliation{%
  \institution{The University of Osaka}
  \city{Osaka}
  \country{Japan}
}
\email{okafuji\_yuki\_xd@cyberagent.co.jp}

\author{Koji Inoue}
\orcid{0000-0002-2929-2559}
\affiliation{%
  \department{Graduate School of Informatics}
  \institution{Kyoto University}
  \city{Kyoto}
  \country{Japan}
}
\affiliation{%
  \institution{The University of Osaka}
  \city{Osaka}
  \country{Japan}
}
\email{inoue@sap.ist.i.kyoto-u.ac.jp}

\author{Yoshiki Ohira}
\orcid{0009-0000-0625-7745}
\affiliation{%
  \institution{CyberAgent}
  \city{Tokyo}
  \country{Japan}
}
\affiliation{%
  \institution{The University of Osaka}
  \city{Osaka}
  \country{Japan}
}
\email{ohira\_yoshiki@cyberagent.co.jp}

%%
%% By default, the full list of authors will be used in the page
%% headers. Often, this list is too long, and will overlap
%% other information printed in the page headers. This command allows
%% the author to define a more concise list
%% of authors' names for this purpose.
\renewcommand{\shortauthors}{Okafuji et al.}

%%
%% The abstract is a short summary of the work to be presented in the
%% article.
\begin{abstract}
Large language model (LLM)-based dialogue systems suffer response delays because generation begins only after final speech recognition. While fixed fillers are a workaround, they become unnatural over time. We propose a two-stage incremental framework that decouples prefatory-response preparation from speech onset. Once user intent becomes predictable, an intent readiness detector triggers LLM-based generation of a short prefatory response. Concurrently, a voice activity projection (VAP) model determines when to deliver it. Through a field experiment with a route-guidance robot in a shopping mall, we evaluated three conditions: no-filler, fixed-filler, and contextual-preface. Both fixed-filler and contextual-preface significantly reduced initial response latency relative to no-filler. Relative to fixed-filler, contextual-preface had significantly longer initial response latency but a significantly shorter initial-to-main gap. Exploratory ratings showed no significant differences. These results indicate a timing trade-off.
\end{abstract}

% Replace the placeholder DOI, ISBN, date, and location with the ACM-provided
% camera-ready metadata after acceptance.



%% A "teaser" image appears between the author and affiliation
%% information and the body of the document, and typically spans the
%% page.
\begin{teaserfigure}
\centering
\includegraphics[width=160mm]{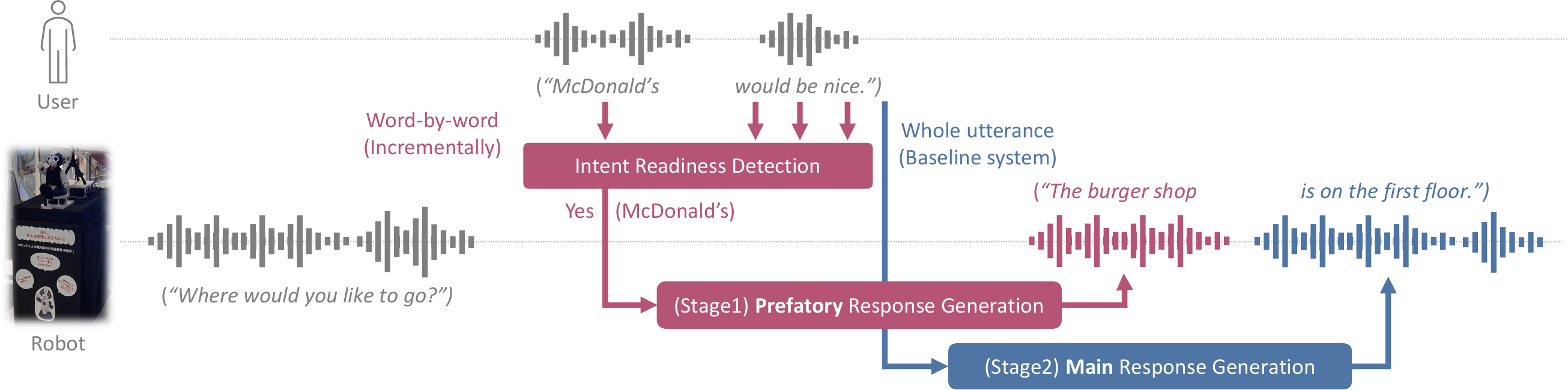}
 \caption{Proposed two-stage architecture for incremental response generation. The system prepares a prefatory response (Stage 1) once user intent becomes predictable. In parallel, the main response (Stage 2) is delivered after the prefatory response.}
 \Description{A two-stage dialogue architecture. Incremental words from a user's utterance enter an intent readiness detector. Once the intent is ready, Stage 1 generates and speaks a short contextual preface. The completed utterance enters the baseline path, where Stage 2 generates and speaks the main response. The diagram illustrates the user request ``McDonald's would be nice,'' followed by the preface ``The burger shop'' and the main response ``is on the first floor.''}
 \label{fig:system}
\end{teaserfigure}

%%
%% This command processes the author and affiliation and title
%% information and builds the first part of the formatted document.
\maketitle

\section{Introduction}

% \begin{figure*}[!t]
%     \centering
%     \includegraphics[width=0.85\linewidth]{figs/systems.png}
%     \caption{Proposed dialogue architecture for two-stage incremental response generation. The system first prepares a prefatory response (Stage 1) once the user’s intent becomes predictable. A VAP model then determines when this prepared response should be spoken. In parallel, the main response (Stage 2) is generated and delivered after the prefatory response.}
%     \label{fig:system}
% \end{figure*}

The rapid advancement of large language models (LLMs) has improved the flexibility and linguistic capability of dialogue systems~\cite{zhang2023large,esteban2024using}.
In practical spoken interaction, however, many systems still rely on cascaded ASR (Automatic Speech Recognition)--LLM--TTS (Text-to-Speech) pipelines because they offer clearer control over safety, role behavior, and external tools than emerging end-to-end or full-duplex systems~\cite{defossez2024moshi,roy2026personaplex,yang2026duplexcascade}.
A major challenge in these cascaded architectures is turn-taking: determining not just what to say, but when to say it~\cite{skantze2021review}.
While humans seamlessly coordinate this process with minimal overlap~\cite{levinson2015timing}, replicating it in machines requires predictive models.
Building on work on incremental dialogue processing, which argues for processing partial input as it becomes available~\cite{schlangen2009general}, Voice Activity Projection (VAP)~\cite{ekstedt2022vap,inoue2024realtime} has gained attention as a robust framework for anticipating upcoming speech activity, advancing beyond earlier turn-transition models~\cite{roddy2018investigating,ekstedt2020turngpt}.

Turn-taking models like VAP have recently been applied to human-robot interaction (HRI)~\cite{skantze2025applying,inoue2025noise}. Yet, fast turn-taking alone is insufficient; a system cannot promptly take the floor if the LLM has not yet prepared a response.
Since naive incremental generation often degrades response consistency~\cite{chiba2025investigating}, an inherent trade-off exists between response speed and generation quality.
To bridge this gap, systems often use conversational fillers to manage delays and maintain interaction flow~\cite{devault2009can,nakanishi2018generating, lala19_interspeech}.
Because generic or repeated fillers can sound unnatural over time~\cite{maslych2025mitigating}, contextually appropriate preliminary responses are preferred~\cite{boukaram2021mitigating}.
Thus, systems need context-aware preliminary responses that can hold the floor while the main response is being formulated.

To address this challenge, we propose a two-stage incremental response framework (Figure~\ref{fig:system}).
This framework prepares a context-aware prefatory response as soon as the user's intent becomes predictable, using turn-taking prediction to determine exactly when to deliver it.
Concurrently, the main response is generated to reduce the delay before the substantive answer, although semantic and prosodic continuity between the prefatory and main responses is not guaranteed.
While previous studies have explored prefetching entire dialogue responses~\cite{mori2025dialogue,ohagi2024investigation}, our approach differs by explicitly separating response preparation from speech onset and focusing solely on an initial contextual prefatory response.

We evaluated this framework on an LLM-based dialogue robot in a shopping mall. Both fixed-filler and contextual-preface significantly reduced initial response latency versus no-filler. Relative to fixed-filler, contextual-preface had significantly longer initial response latency but a significantly shorter initial-to-main gap.

\section{Proposed System}

This section describes our two-stage incremental response framework, covering the overall architecture, the intent readiness detector, and its integration into a dialogue robot.

\subsection{Design and Architecture}
Our proposed system extends a VAP-based dialogue system~\cite{inoue2025noise} by adding Intent Readiness Detection and prefatory-response generation modules (Figure~\ref{fig:system}). Incremental Speech-to-Text (STT) continuously updates a partial recognition result, and the intent readiness detector estimates whether the user's intent has become predictable enough to prepare a short prefatory response. Once the detector fires, the system generates that prefatory response from the dialogue history and partial utterance. VAP~\cite{ekstedt2022vap,inoue2024realtime} independently estimates when the user is about to yield the turn; if a prefatory response is ready at that point, the system speaks it. Once the full STT result becomes available, the system generates the main response from the completed utterance. In this way, the architecture realizes a two-stage response structure while controlling preparation timing and speech onset separately.

Each system response is therefore divided into two stages: (i) a brief prefatory response that occupies the immediate post-turn silence and (ii) a main response that carries the task-relevant information. Unlike fixed fillers or backchannels used only as waiting feedback~\cite{shiwa2009response, boukaram2021mitigating,kang2024feedback,jeong2019fillers}, the prefatory response is conditioned on the user's emerging intent. For example, if the system asks, \textit{``Where would you like to go?''} and the user replies, \textit{``McDonald's would be nice,''} the destination can already be inferred once the partial utterance reaches ``\textit{McDonald's}.'' The system can then prepare a prefatory response such as \textit{``The burger shop, \ldots''} before the user's utterance ends and deliver the main response after turn completion.

The main response is generated while the prefatory response is being spoken, reducing the delay before the substantive answer. To limit the risk of early prediction errors, the prefatory response must be no longer than 10 Japanese characters, must not introduce new information or state facts definitively, must not express strong agreement or make concrete suggestions, and must remain limited to filler-like or bridging expressions.

\subsection{Intent Readiness Detection Model}

In this study, Intent Readiness Detection is formulated as a binary classification problem that determines whether the user's intent has become sufficiently predictable from the current utterance prefix, conditioned on the immediately preceding system utterance. Here, intent refers to the emerging direction of the user's utterance rather than a predefined intent class, and the detector predicts only whether this intent is ready or not ready for prefatory-response preparation. At each incremental STT update, the detector receives the previous system utterance and the current user prefix, then outputs an intent-readiness score indicating whether it is safe to begin preparing the prefatory response. During inference, prefatory-response generation is triggered when this score exceeds the deployment threshold.

We trained the detector in two stages: first on an LLM-pseudo-labeled corpus (764,976 instances) and then by fine-tuning on a human-annotated corpus (30,740 instances). Both corpora were diversified with an LLM to include hesitations, self-repairs, ellipses, and ambiguous replies.

The classifier was built on a Japanese ModernBERT 30M encoder initialized from a publicly available pretrained checkpoint~\cite{warner2024modernbert}. The previous system utterance and current user prefix were concatenated into a single input sequence and fed to a binary classification head. Class imbalance was addressed using focal loss and weighted sampling.
The decision threshold was empirically set to 0.35.

In offline evaluation, the final model achieved 94.31\% accuracy and 0.89 macro F1 on the held-out test split. The goal of the detector is not to reconstruct the final meaning of the user's utterance, but to estimate whether enough information is available to prepare a prefatory response. Accordingly, when later information may reverse the intended meaning, the detector should behave conservatively and delay the trigger. As shown later in the field experiment, some utterances become intent-ready only after the full utterance has been observed.

\subsection{Dialogue Robot System}

We implemented the proposed method in a dialogue robot system for guidance in commercial facilities. The system consists of a small humanoid robot, a microphone input, a knowledge base containing information for the facility, and external APIs for speech recognition and response generation. The robot assists users by providing destination and facility information, generating responses on the basis of the dialogue history.

For STT, we used the Google Speech-to-Text API and exploited both incremental and final recognition results. Turn-end prediction was handled by a locally running VAP model~\cite{inoue2025noise}, with fallback to the final STT result when the confidence of the turn-taking prediction was insufficient. This yielded a hybrid policy that aims for early turn entry when acoustically reliable while waiting for recognition finalization in uncertain cases. For prefatory-response generation, we used the OpenAI API model \texttt{gpt-4o-mini}, and for main-response generation, \texttt{gpt-4o}. The former emphasizes low-latency generation of short prefatory responses, whereas the latter generates substantively appropriate responses grounded in facility knowledge. The generated text was synthesized by the Japanese TTS system VOICEVOX and output as the robot's speech. Because our target is real-world deployment in Japanese facilities, all processing in this study was conducted in Japanese.

\section{Field Experiment}

To evaluate the effectiveness of the proposed framework, we conducted a real-world field experiment using a route-guidance robot.

\subsection{Settings}

\begin{figure}[!t]
    \centering
    \includegraphics[width=0.85\columnwidth]{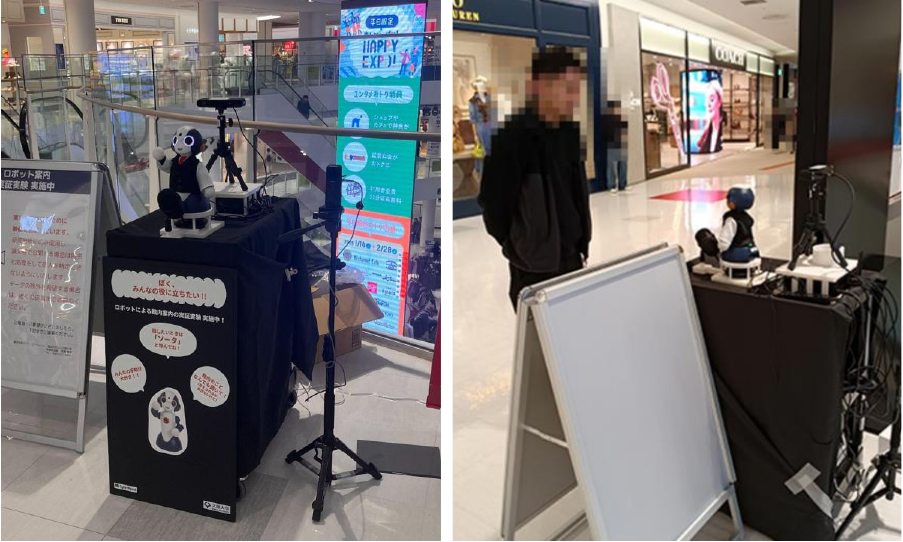}
    \caption{Example of users interacting with robots.}
    \Description{Two photographs of the field deployment in a shopping mall. The left photograph shows a small humanoid guidance robot mounted on a kiosk-like stand together with microphones, cameras, signs, and other equipment. The right photograph shows a visitor standing in front of and interacting with the robot in the open mall space.}
    \label{fig:experiment}
\end{figure}

In the field experiment, we used a guiding robot and deployed the robot in an open space within a shopping mall in Japan. Visitors freely interacted with the robot without any specific given dialogue task. This situation is important for evaluating the robustness of the proposed method because it reflects real-world uncertainty, including environmental noise, diverse user intentions, hesitations, and mid-utterance repairs. Figure~\ref{fig:experiment} shows an example of a user interacting with a robot.

The field study was conducted on three weekdays in January 2026. The study was approved by the ethics review board of The University of Osaka. Signs at the deployment site explained the purpose and procedure of the study so that visitors could decide for themselves whether to interact with the robot.

The field experiment used three conditions, which were deployed at randomly assigned times: (i) a \textit{no-filler condition}, in which the robot produces only the main response after the user's utterance ended; (ii) a \textit{fixed-filler condition}, in which the robot first utters a fixed filler, such as ``yeah'' or ``I see,'' while preparing the main response; and (iii) a \textit{contextual-preface condition}, in which the robot first prepares a short contextual prefatory response from the user's emerging intent and then delivers the main response after it becomes available. In all conditions, the timing of system turn entry at the end of the user's turn was controlled by the same VAP-based turn-taking mechanism.

We evaluated interactions that contained at least two turns. We collected 47 interactions (205 response instances) in the \textit{no-filler condition}, 52 interactions (188 response instances) in the \textit{fixed-filler condition}, and 75 interactions (251 response instances) in the \textit{contextual-preface condition}, where those users are different people.
%We also collected 30 post-interaction questionnaires per condition, finding no significant differences across conditions. Because conversational fillers and prefatory responses are peripheral elements of dialogue, it is inherently difficult to observe significant differences in subjective evaluations among such conditions. Therefore, we emphasize that the system successfully maintained (did not degrade) users' subjective impressions even when attempting challenging early contextual prefaces, while achieving the crucial benefit of objectively shortening actual response times.

% \begin{figure*}[!t]
%     \centering
%     \begin{subfigure}[t]{0.42\linewidth}
%         \centering
%         \includegraphics[width=\linewidth]{figs/first-response_response_distribution.png}
%         \caption{Distribution of initial response latency}
%         \label{fig:first-response}
%     \end{subfigure}
%     \hfill
%     \begin{subfigure}[t]{0.42\linewidth}
%         \centering
%         \includegraphics[width=\linewidth]{figs/second-response_response_distribution.png}
%         \caption{Distribution of initial-to-main gap}
%         \label{fig:second-response}
%     \end{subfigure}
%     \caption{Distribution of response latency metrics in the field experiment.}
%     \label{fig:response-distributions}
% \end{figure*}

\subsection{Evaluation Metrics}

For log analysis, we used the following objective metrics:
\begin{itemize}
    \item \textbf{Initial response latency}: the time from the end of the user's utterance to the robot's first response. In the \textit{no-filler condition}, this corresponds to the onset of the main response; in the \textit{fixed-filler} and \textit{contextual-preface conditions}, it corresponds to the onset of the initial response. This metric captures response-onset timing in the system logs, but excludes the delay before the STT result appears.
    \item \textbf{Initial-to-main gap}: the time from the end of the initial response to the beginning of the main response. This metric is defined only for the \textit{fixed-filler} and \textit{contextual-preface conditions}.
    \item \textbf{Trigger position}: in the \textit{contextual-preface condition}, the point at which preparation of the prefatory response began, expressed as the character-level progress ratio relative to the final user-utterance length.
\end{itemize}
Because initial response latency and initial-to-main gap exhibited skewed distributions, we used Kruskal--Wallis tests for omnibus comparisons and Mann--Whitney $U$ tests for pairwise comparisons.

For breakdown analysis, all authors annotated prefatory responses in the \textit{contextual-preface condition} using four categories~\citet{higashinaka2021integrated}: FRAGMENT (incomplete or cut-off output), GENERIC QUESTION (vague question after a specific request), SLOT MISMATCH (wrong slot or destination type), and EXPECTATION VIOLATION (ill-fitting response).

We additionally collected 30 post-interaction questionnaires per condition. Participants rated conversational pacing, naturalness of the initial response, appropriateness of the full response, and overall satisfaction on 7-point Likert scales.
%The questionnaire analysis was treated as exploratory and was intended to examine whether the early-preface strategy produced detectable changes in perceived interaction quality.

\subsection{Results}

\paragraph{Response Latency}

\begin{figure}[!t]
    \centering
        \includegraphics[width=0.85\columnwidth]{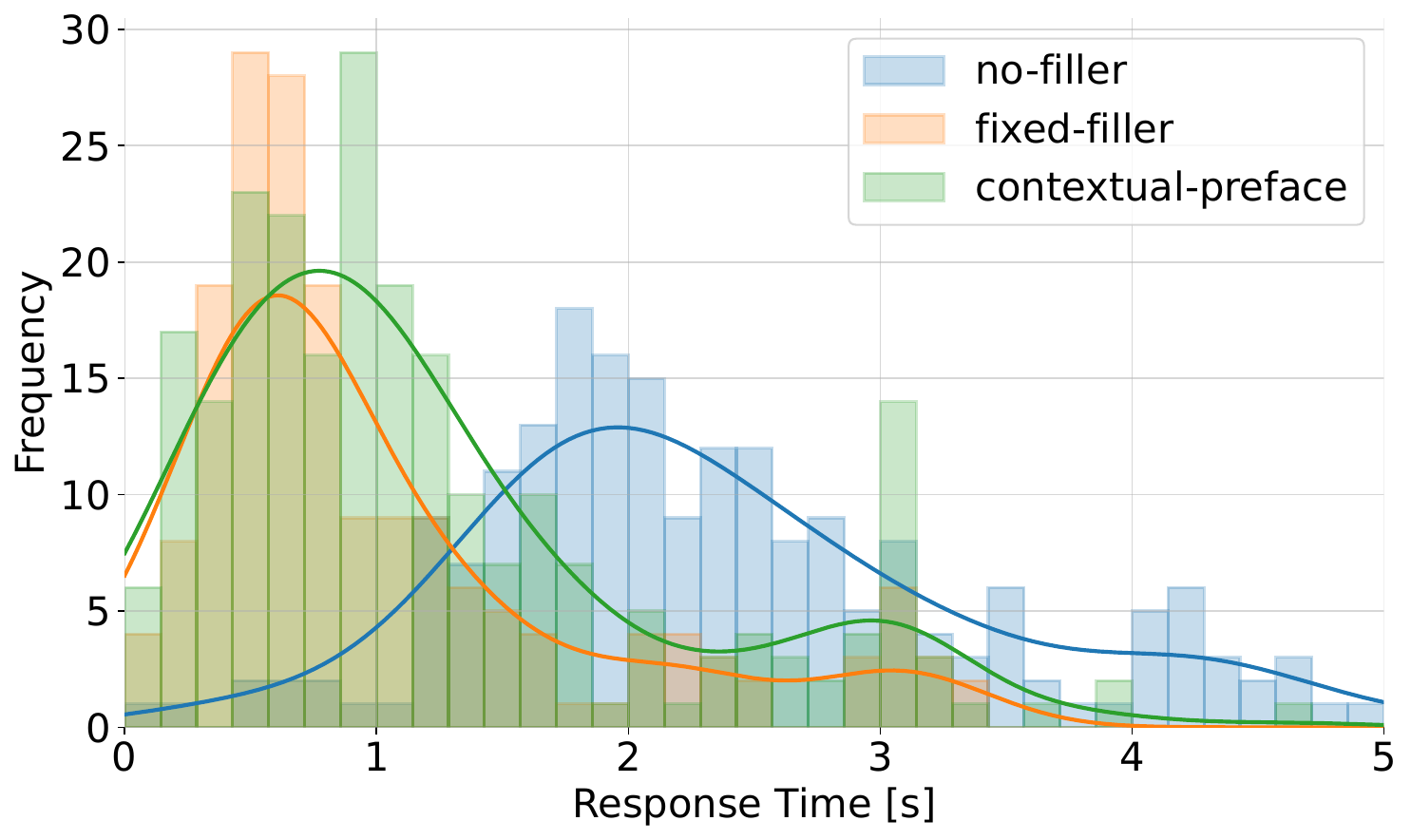}
        \caption{Distribution of initial response latency}
        \Description{Overlaid histograms and density curves of initial response latency, ranging from zero to five seconds, for the no-filler, fixed-filler, and contextual-preface conditions. The no-filler distribution is shifted toward longer response times and peaks at approximately two seconds. Both the fixed-filler and contextual-preface distributions are concentrated near one second, with fixed-filler showing slightly shorter initial latencies than contextual-preface.}
        \label{fig:first-response}
\end{figure}

\begin{figure}[!t]
    \centering
        \includegraphics[width=0.85\columnwidth]{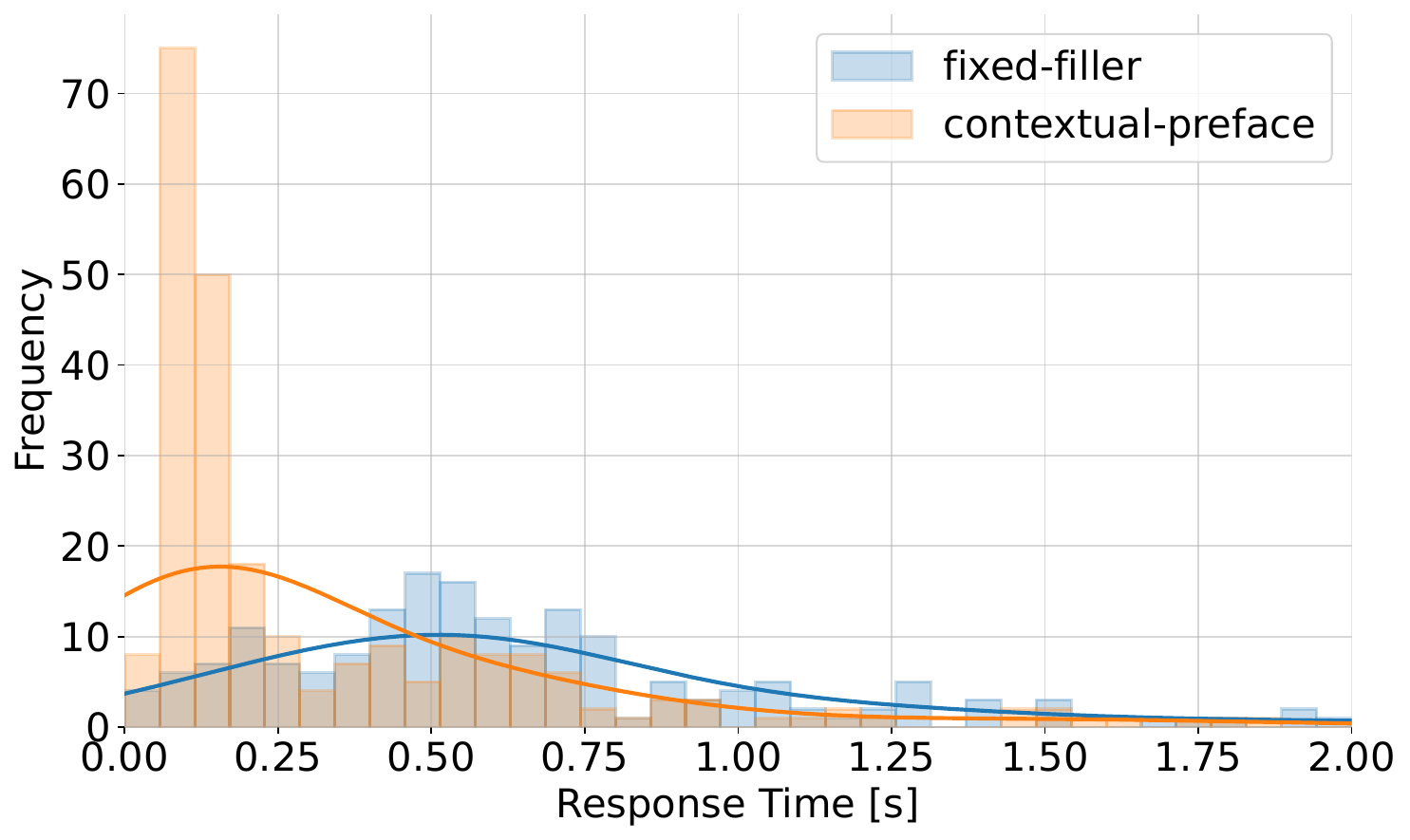}
        \caption{Distribution of initial-to-main gap}
        \Description{Overlaid histograms and density curves of the time gap between the end of the initial response and the beginning of the main response, ranging from zero to two seconds, for the fixed-filler and contextual-preface conditions. The contextual-preface distribution is strongly concentrated near zero, whereas the fixed-filler distribution is broader and extends toward longer delays, indicating a shorter initial-to-main gap for contextual-preface.}
        \label{fig:second-response}
\end{figure}

Table~\ref{tab:field_latency_detail} summarizes the latency results and Figure~\ref{fig:first-response} shows the distribution of response latency. Initial response latency differed significantly across conditions ($H = 211.40$, $p < .001$). Both \textit{fixed-filler} and \textit{contextual-preface} were significantly faster than \textit{no-filler}, and \textit{fixed-filler} was also faster than \textit{contextual-preface} (Bonferroni-corrected $p = .027$).
However, the initial-to-main gap was significantly shorter in \textit{contextual-preface} than in \textit{fixed-filler} ($U = 35139.0$, $p < .001$) as depicted in Figure~\ref{fig:second-response}.

\begin{table}[!t]
    \centering
    \caption{Overall response latency in the field experiment. The numerical value of latencies represents Mean (Median).}
    \label{tab:field_latency_detail}
    \small
    \setlength{\tabcolsep}{4pt}
    \begin{tabular}{@{}lccc@{}}
    \toprule
    \multicolumn{1}{c}{Condition} & \# resp. & Initial latency [s] & Initial-to-main gap [s] \\
    \midrule
    \textit{no-filler} & 205 & 2.45 (2.19) & --- \\
    \textit{fixed-filler} & 188 & 0.94 (0.70) & 0.74 (0.57) \\
    \textit{contextual-preface} & 251 & 1.15 (0.92) & 0.43 (0.16) \\
    \bottomrule
    \end{tabular}
\end{table}

Separating turn-entry sources into VAP and STT further clarified the source of the latency differences. VAP-triggered onsets accounted for 73.2\%, 78.7\%, and 77.7\% of responses in the \textit{no-filler}, \textit{fixed-filler}, and \textit{contextual-preface} conditions, respectively. The corresponding mean initial response latencies for VAP-triggered onsets were 2.20, 0.68, and 0.92~s, whereas those for STT-triggered onsets were 3.13, 1.89, and 1.96~s. This pattern suggests that the shorter initial-to-main gap observed in the \textit{contextual-preface condition} was not due to more frequent VAP triggering. It is instead consistent with greater temporal overlap between main generation and the initial response by Intent Readiness Detection.

\paragraph{Timing of Intent Readiness Detection}
\label{sec:field_complete_timing}

To analyze the proposed method more finely, we also examined when the intent readiness detector fired for the 251 utterances in the \textit{contextual-preface condition}. The detector fired after a mean of 5.53 characters, and the mean final utterance length was 9.89 characters. The mean per-utterance normalized trigger position was 69.2\%. The detector triggered before the end of the user utterance in 58.2\% of cases, indicating that the system could often begin preparing a prefatory response before the user had finished speaking, while still waiting for the full utterance when early prediction was less reliable.

\paragraph{Dialogue Breakdowns}
Twenty of the 251 prefatory responses in the \textit{contextual-preface condition} were annotated as dialogue breakdowns. The most frequent failure types were FRAGMENT responses (10 cases, 50.0\%) and GENERIC QUESTION responses (8 cases, 40.0\%), showing that both output truncation and weak contextual grounding were major failure modes. For example, in a GENERIC QUESTION case, when the user asked \textit{``Where is UNIQLO?''}, the robot responded \textit{``Are you looking for something?''}, ignoring the specific request. By contrast, SLOT MISMATCH and EXPECTATION VIOLATION responses each occurred only once (1 case, 5.0\%), suggesting that these error types were comparatively rare.

\paragraph{Subjective Impressions}
%Kruskal-Wallis tests revealed no significant differences across conditions for any subjective metric ($p > .05$). Because peripheral elements like prefaces rarely alter overall subjective evaluations, these results successfully confirm that our early contextual prefaces maintained user impressions without degradation while objectively reducing latency.
Kruskal--Wallis tests detected no statistically significant differences across conditions for any of the four questionnaire items (all $p > .05$). Given the exploratory sample of 30 questionnaires per condition, these null results should not be interpreted as evidence of equivalence or as demonstrating the absence of subjective effects.

\section{Discussion and Conclusion}
The field experiments demonstrated that using context-aware prefatory responses can fill part of the post-turn waiting time, thereby enhancing the practicality of response generation for robots deployed in real-world settings. Unlike fixed fillers, the proposed method not only masked latency at turn onset; it also shortened the gap before the substantive response, while maintaining substantially faster initial response timing than the no-filler baseline.

No statistically significant differences were detected in the four exploratory subjective ratings. Given the exploratory sample of 30 questionnaires per condition, these null results do not establish either improvement or equivalence in perceived interaction quality. Larger targeted studies are needed to determine whether contextual prefaces affect overall user experience.

The trigger-position and breakdown analyses also clarify the trade-off introduced by early preparation. Selective triggering is necessary to avoid acting on insufficient partial input, while the observed fragment and generic-question failures indicate that stable, contextually grounded generation of short prefaces remains a key bottleneck. Because the prefatory and main responses are generated independently, improving their semantic and prosodic continuity is another important direction for future work.

%Crucially, while the primary benefit observed was in objective timing, this speedup did not compromise the user experience. Because prefaces are peripheral conversational elements, subjective evaluation scores remained stable across conditions. This stability confirms that our system can perform challenging early prefaces without degrading perceived naturalness. The trigger-position and breakdown analyses highlight the trade-off of this approach. Selective early preparation is necessary because early prediction can introduce semantic or prosodic discontinuity when the prefatory response and main response are generated independently, and the remaining bottleneck is the stability of short-response generation under partial input.

This study is limited to Japanese route-guidance interactions in one shopping mall, and the observed latencies depend on the specific ASR, LLM, and TTS components used in our implementation. Because the prefatory and main responses are generated independently, future work should improve their semantic and prosodic continuity, strengthen contextual grounding and output completeness, support user barge-in, and test the framework across broader domains and languages.

\section*{Safe and Responsible Innovation Statement}

This study received institutional ethics approval. In the shopping mall, clear signs informed visitors and allowed opt-outs, while all collected data was anonymized to protect privacy. To mitigate LLM hallucinations and inappropriate outputs in public, prefatory responses were strictly limited to short, non-committal expressions (under 10 characters). By avoiding definitive statements, we ensured a safe, controllable, and responsible guidance system.

%\section*{Acknowledgment}
\begin{acks}
This work was supported by JST PRESTO (JPMJAX2103) and JST BOOST (JPMJBY24A7).
\end{acks}

\bibliographystyle{ACM-Reference-Format}
\bibliography{custom}

@article{levinson2015timing,
  author = {Levinson, Stephen C. and Torreira, Francisco},
  title = {Timing in turn-taking and its implications for processing models of language},
  journal = {Frontiers in Psychology},
  volume = {6},
  pages = {731},
  year = {2015},
  doi = {10.3389/fpsyg.2015.00731}
}

@article{skantze2021review,
  author = {Skantze, Gabriel},
  title = {Turn-taking in conversational systems and human-robot interaction: A review},
  journal = {Computer Speech \& Language},
  volume = {67},
  pages = {101178},
  year = {2021},
  doi = {10.1016/j.csl.2020.101178}
}

@inproceedings{schlangen2009general,
  author = {Schlangen, David and Skantze, Gabriel},
  title = {A General, Abstract Model of Incremental Dialogue Processing},
  booktitle = {Proceedings of the 12th Conference of the European Chapter of the ACL (EACL 2009)},
  pages = {710--718},
  address = {Athens, Greece},
  publisher = {Association for Computational Linguistics},
  year = {2009},
  url = {https://aclanthology.org/E09-1081/}
}

@inproceedings{ekstedt2022vap,
  author = {Ekstedt, Erik and Skantze, Gabriel},
  title = {Voice Activity Projection: Self-supervised Learning of Turn-taking Events},
  booktitle = {Interspeech 2022},
  pages = {5190--5194},
  year = {2022},
  doi = {10.21437/Interspeech.2022-10955}
}

@article{inoue2024realtime,
  author = {Inoue, Koji and Jiang, Bing'er and Ekstedt, Erik and Kawahara, Tatsuya and Skantze, Gabriel},
  title = {Real-time and Continuous Turn-taking Prediction Using Voice Activity Projection},
  journal = {arXiv preprint arXiv:2401.04868},
  year = {2024},
  note = {Accepted to the 14th International Workshop on Spoken Dialogue Systems Technology (IWSDS 2024)},
  doi = {10.48550/arXiv.2401.04868},
  url = {https://arxiv.org/abs/2401.04868}
}

@INPROCEEDINGS{inoue2025noise,
  author={Inoue, Koji and Okafuji, Yuki and Baba, Jun and Ohira, Yoshiki and Hyodo, Katsuya and Kawahara, Tatsuya},
  booktitle={2025 IEEE/RSJ International Conference on Intelligent Robots and Systems (IROS)}, 
  title={A Noise-Robust Turn-Taking System for Real-World Dialogue Robots: A Field Experiment}, 
  year={2025},
  volume={},
  number={},
  pages={874-879},
  doi={10.1109/IROS60139.2025.11246533}
}

@inproceedings{higashinaka2021integrated,
  author = {Higashinaka, Ryuichiro and Araki, Masahiro and Tsukahara, Hiroshi and Mizukami, Masahiro},
  title = {Integrated taxonomy of errors in chat-oriented dialogue systems},
  booktitle = {Annual Meeting of the Special Interest Group on Discourse and Dialogue (SIGDIAL)},
  pages = {89--98},
  year = {2021}
}

@article{warner2024modernbert,
  author = {Warner, Benjamin and Chaffin, Antoine and Clavi{\'e}, Benjamin and Weller, Orion and Hallstr{\"o}m, Oskar and Taghadouini, Said and Gallagher, Alexis and Biswas, Raja and Ladhak, Faisal and Aarsen, Tom and Cooper, Nathan and Adams, Griffin and Howard, Jeremy and Poli, Iacopo},
  title = {Smarter, Better, Faster, Longer: A Modern Bidirectional Encoder for Fast, Memory Efficient, and Long Context Finetuning and Inference},
  journal = {arXiv preprint arXiv:2412.13663},
  year = {2024},
  doi = {10.48550/arXiv.2412.13663},
  url = {https://arxiv.org/abs/2412.13663}
}

@article{shiwa2009response,
  author  = {Shiwa, Toshiyuki and Kanda, Takayuki and Imai, Michita and Ishiguro, Hiroshi and Hagita, Norihiro},
  title   = {How Quickly Should a Communication Robot Respond? Delaying Strategies and Habituation Effects},
  journal = {International Journal of Social Robotics},
  volume  = {1},
  number  = {2},
  pages   = {141--155},
  year    = {2009},
  doi     = {10.1007/s12369-009-0012-8}
}

@article{kang2024feedback,
  author  = {Kang, Dahyun and Nam, Changjoo and Kwak, Sonya S.},
  title   = {Robot Feedback Design for Response Delay},
  journal = {International Journal of Social Robotics},
  volume  = {16},
  pages   = {341--361},
  year    = {2024},
  doi     = {10.1007/s12369-023-01068-z}
}

@inproceedings{jeong2019fillers,
  author    = {Jeong, Yuin and Lee, Juho and Kang, Younah},
  title     = {Exploring Effects of Conversational Fillers on User Perception of Conversational Agents},
  booktitle = {Extended Abstracts of the 2019 CHI Conference on Human Factors in Computing Systems},
  articleno = {LBW2715},
  pages     = {1--6},
  year      = {2019},
  publisher = {Association for Computing Machinery},
  doi       = {10.1145/3290607.3312913}
}

@article{zhang2023large,
    title = {Large language models for human–robot interaction: A review},
    journal = {Biomimetic Intelligence and Robotics},
    volume = {3},
    number = {4},
    pages = {100131},
    year = {2023},
    issn = {2667-3797},
    doi = {https://doi.org/10.1016/j.birob.2023.100131},
    url = {https://www.sciencedirect.com/science/article/pii/S2667379723000451},
    author = {Ceng Zhang and Junxin Chen and Jiatong Li and Yanhong Peng and Zebing Mao}
}

@InProceedings{esteban2024using,
    author="Esteban-Lozano, Iv{\'a}n
    and Castro-Gonz{\'a}lez, {\'A}lvaro
    and Mart{\'i}nez, Paloma",
    editor="Degen, Helmut
    and Ntoa, Stavroula",
    title="Using a LLM-Based Conversational Agent in the Social Robot Mini",
    booktitle="Artificial Intelligence in HCI",
    year="2024",
    publisher="Springer Nature Switzerland",
    address="Cham",
    pages="15--26",
    isbn="978-3-031-60615-1",
    url="https://doi.org/10.1007/978-3-031-60615-1_2"
}

@article{defossez2024moshi,
  title={Moshi: a speech-text foundation model for real-time dialogue},
  author={D{\'e}fossez, Alexandre and Mazar{\'e}, Laurent and Orsini, Manu and Royer, Am{\'e}lie and P{\'e}rez, Patrick and J{\'e}gou, Herv{\'e} and Grave, Edouard and Zeghidour, Neil},
  journal={arXiv preprint arXiv:2410.00037},
  year={2024},
  url = {https://doi.org/10.48550/arXiv.2410.00037},
}

@article{roy2026personaplex,
  title={Persona{P}lex: {V}oice and Role Control for Full Duplex Conversational Speech Models},
  author={Roy, Rajarshi and Raiman, Jonathan and Lee, Sang-gil and Ene, Teodor-Dumitru and Kirby, Robert and Kim, Sungwon and Kim, Jaehyeon and Catanzaro, Bryan},
  journal={arXiv preprint arXiv:2602.06053},
  year={2026},
  url={https://doi.org/10.48550/arXiv.2602.06053}
}

@article{yang2026duplexcascade,
  title={DuplexCascade: {F}ull-Duplex Speech-to-Speech Dialogue with VAD-Free Cascaded ASR-LLM-TTS Pipeline and Micro-Turn Optimization},
  author={Yang, Jianing and Fujita, Yusuke and Sudo, Yui},
  journal={arXiv preprint arXiv:2603.09180},
  year={2026},
  url={https://doi.org/10.48550/arXiv.2603.09180}
}

@inproceedings{roddy2018investigating,
  title     = {Investigating Speech Features for Continuous Turn-Taking Prediction Using LSTMs},
  author    = {Matthew Roddy and Gabriel Skantze and Naomi Harte},
  year      = {2018},
  booktitle = {{Interspeech 2018}},
  pages     = {586--590},
  doi       = {10.21437/Interspeech.2018-2124},
  issn      = {2958-1796},
}

@inproceedings{ekstedt2020turngpt,
    title = "{T}urn{GPT}: a Transformer-based Language Model for Predicting Turn-taking in Spoken Dialog",
    author = "Ekstedt, Erik  and
      Skantze, Gabriel",
    editor = "Cohn, Trevor  and
      He, Yulan  and
      Liu, Yang",
    booktitle = "Findings of the Association for Computational Linguistics: EMNLP 2020",
    month = nov,
    year = "2020",
    address = "Online",
    publisher = "Association for Computational Linguistics",
    url = "https://aclanthology.org/2020.findings-emnlp.268/",
    doi = "10.18653/v1/2020.findings-emnlp.268",
    pages = "2981--2990",
}

@inproceedings{skantze2025applying,
    author = {Skantze, Gabriel and Irfan, Bahar},
    title = {Applying General Turn-taking Models to Conversational Human-Robot Interaction},
    year = {2025},
    publisher = {IEEE Press},
    booktitle = {Proceedings of the 2025 ACM/IEEE International Conference on Human-Robot Interaction},
    pages = {859–868},
    numpages = {10},
    location = {Melbourne, Australia},
    series = {HRI '25},
    url = {https://dl.acm.org/doi/10.5555/3721488.3721593}
}

@inproceedings{chiba2025investigating,
    title = "Investigating the Impact of Incremental Processing and Voice Activity Projection on Spoken Dialogue Systems",
    author = "Chiba, Yuya  and
      Higashinaka, Ryuichiro",
    editor = "Rambow, Owen  and
      Wanner, Leo  and
      Apidianaki, Marianna  and
      Al-Khalifa, Hend  and
      Eugenio, Barbara Di  and
      Schockaert, Steven",
    booktitle = "Proceedings of the 31st International Conference on Computational Linguistics",
    month = jan,
    year = "2025",
    address = "Abu Dhabi, UAE",
    publisher = "Association for Computational Linguistics",
    url = "https://aclanthology.org/2025.coling-main.249/",
    pages = "3687--3696",
}

@inproceedings{nakanishi2018generating,
  title={Generating Fillers Based on Dialog Act Pairs for Smooth Turn-Taking by Humanoid Robot},
  author={Ryosuke Nakanishi and Koji Inoue and Shizuka Nakamura and Katsuya Takanashi and Tatsuya Kawahara},
  booktitle={International Workshop on Spoken Dialogue Systems Technology},
  year={2018},
  url={https://api.semanticscholar.org/CorpusID:52572791}
}

@inproceedings{boukaram2021mitigating,
    author = {Boukaram, Halim-Antoine and Ziadee, Micheline and Sakr, Majd F},
    title = {Mitigating the Effects of Delayed Virtual Agent Response Time Using Conversational Fillers},
    year = {2021},
    isbn = {9781450386203},
    publisher = {Association for Computing Machinery},
    address = {New York, NY, USA},
    url = {https://doi.org/10.1145/3472307.3484181},
    doi = {10.1145/3472307.3484181},
    booktitle = {Proceedings of the 9th International Conference on Human-Agent Interaction},
    pages = {130–138},
    numpages = {9},
    location = {Virtual Event, Japan},
    series = {HAI '21}
}

@inproceedings{maslych2025mitigating,
    author = {Maslych, Mykola and Katebi, Mohammadreza and Lee, Christopher and Hmaiti, Yahya and Ghasemaghaei, Amirpouya and Pumarada, Christian and Palmer, Janneese and Segarra Martinez, Esteban and Emporio, Marco and Snipes, Warren and McMahan, Ryan P. and LaViola Jr., Joseph J.},
    title = {Mitigating Response Delays in Free-Form Conversations with LLM-powered Intelligent Virtual Agents},
    year = {2025},
    isbn = {9798400715273},
    publisher = {Association for Computing Machinery},
    address = {New York, NY, USA},
    url = {https://doi.org/10.1145/3719160.3736636},
    doi = {10.1145/3719160.3736636},
    booktitle = {ACM Conference on Conversational User Interfaces (CUI)},
    articleno = {49},
    numpages = {15},
}

@inproceedings{ohagi2024investigation,
  title     = {Investigation of look-ahead techniques to improve response time in spoken dialogue system},
  author    = {Masaya Ohagi and Tomoya Mizumoto and Katsumasa Yoshikawa},
  year      = {2024},
  booktitle = {INTERSPEECH},
  pages     = {3580--3584}
}

@inproceedings{mori2025dialogue,
  title     = {Dialogue Response Prefetching Based on Semantic Similarity and Prediction Confidence of Language Model},
  author    = {Kiyotada Mori and Seiya Kawano and Angel García Contreras and Koichiro Yoshino},
  year      = {2025},
  booktitle = {INTERSPEECH},
  pages     = {3045--3049}
}

@inproceedings{lala19_interspeech,
  title     = {{Analysis of Effect and Timing of Fillers in Natural Turn-Taking}},
  author    = {Divesh Lala and Shizuka Nakamura and Tatsuya Kawahara},
  year      = {2019},
  booktitle = {INTERSPEECH},
  pages     = {4175--4179}
}

@inproceedings{devault2009can,
  title={Can I finish? Learning when to respond to incremental interpretation results in interactive dialogue},
  author={DeVault, David and Sagae, Kenji and Traum, David},
  booktitle={Annual Meeting of the Special Interest Group on Discourse and Dialogue (SIGDIAL)},
  pages={11--20},
  year={2009}
}

\end{document}